\documentclass[sn-vancouver]{sn-jnl}

\usepackage{booktabs, multirow} 
\usepackage{soul}
\usepackage{float}
\usepackage{changepage,threeparttable}
\usepackage{adjustbox}
\usepackage{lscape}

\usepackage{caption}

\jyear{2021}%

\theoremstyle{thmstyleone}%
\theoremstyle{thmstyletwo}%

\theoremstyle{thmstylethree}%

\let\oldReturn\Return
\renewcommand{\Return}{\State\oldReturn}
\begin{document}

\title[A Novel Nearest Neighbors Algorithm Based on Power Muirhead Mean]{
A Novel Nearest Neighbors Algorithm Based on Power Muirhead Mean}


\author*[1]{\fnm{Kourosh} \sur{Shahnazari}}\email{kourosh@aut.ac.ir}
\equalcont{}

\author*[1]{\fnm{Seyed Moein} \sur{Ayyoubzadeh}}\email{s.m.ayyoubzadeh@aut.ac.ir}
\equalcont{}

\affil*[1]{\orgdiv\orgname{Amirkabir University of Technology}, \orgaddress{\city{Tehran}, \country{Iran}}}


\abstract{Abstract:
This paper introduces the innovative Power Muirhead Mean K-Nearest Neighbors (PMM-KNN) algorithm, a novel data classification approach that combines the K-Nearest Neighbors method with the adaptive Power Muirhead Mean operator. The proposed methodology aims to address the limitations of traditional KNN by leveraging the Power Muirhead Mean for calculating the local means of K-nearest neighbors in each class to the query sample. Extensive experimentation on diverse benchmark datasets demonstrates the superiority of PMM-KNN over other classification methods. Results indicate statistically significant improvements in accuracy on various datasets, particularly those with complex and high-dimensional distributions. The adaptability of the Power Muirhead Mean empowers PMM-KNN to effectively capture underlying data structures, leading to enhanced accuracy and robustness. The findings highlight the potential of PMM-KNN as a powerful and versatile tool for data classification tasks, encouraging further research to explore its application in real-world scenarios and the automation of Power Muirhead Mean parameters to unleash its full potential.}




\maketitle

\section{Introduction}\label{sec1}

Numerous domains, including pattern recognition and artificial intelligence, statistics, cognitive psychology, vision analysis, and medicine, rely on object classification for their research and practical applications.

The K-Nearest Neighbors(KNN) classifier tackles the problem of classification by first determining the distance between a new sample to be classified and the training samples, then observing the k nearest neighbors for the new sample, and finally deciding whether or not the new sample belongs to the class that shares the most neighbors with the new sample. \citep{fknn}

Typically, KNN is a useful classifier, but it is worth mentioning that, because the parametric form of the density function cannot be assumed, non-parametric classifiers need a large number of samples. When working with a tiny sample size, this might be a serious problem \citep{raudys1991small,cover1967nearest,fukunaga2013introduction}. 

Moreover, a well-known problem with KNN classifiers is that they are susceptible to outliers that distort the class distribution.\citep{fukunaga2013introduction}

The majority voting principle is the foundation of KNN, which determines the class of a new sample by looking at its closest neighbors and the class that constitutes the majority of those neighbors. In the case that a data set is obviously unbalanced, one of the drawbacks of the majority voting principle is that the classified samples of the class or classes that have a large number of samples have a tendency to dominate the prediction of the new sample. This is simply due to the fact that they are frequently more numerous among the k nearest neighbors.\citep{coomans1982alternative}\citep{hajizadeh2022mutual}

The Muirhead Mean (MM), first introduced by Muirhead in 1902, is a generic aggregation function. It can be defined as a function of means and has been used in various applications owing to its ability to acknowledge interrelationships. \citep{liu2018some}

The Power Muirhead Mean (PMM) is an extension of the MM, designed to mitigate the adverse impact of outliers and emphasize central data points more effectively. This is achieved by combining the MM with a Power Average (PA), thus allowing the algorithm to adapt to various data distributions more robustly.

In this research, we offer a method for overcoming the limitations of the KNN classifier by using the Power Muirhead Mean of K nearest neighbors for each class separately.

\section{Literature Review}\label{sec2}
This section summarizes the theoretical foundations of the KNN classifier, Muirhead Mean(MM) Operator, Power Average(PA), and Power Muirhead Mean(PMM).
\subsection{K-Nearest Neighbors Classifier}\label{subsec1}
The classic KNN algorithm is one of the earliest and most basic techniques for classifying patterns. In spite of this, it often produces competitive outcomes, and in particular fields, when supplemented with previous knowledge, it has greatly improved the state of the art. The KNN rule classifies each unlabeled instance according to the majority label of its k-nearest neighbors in the training set. Therefore, its success is highly dependent on the distance measure used to locate the closest neighbors. In the absence of previous information, the majority of KNN classifiers utilize a basic Euclidean metric to quantify dissimilarities across vector input instances. The following equation describes Euclidean distance. \citep{sun2010adaptive}
\begin{equation}
d({V}, {M}) = \sqrt {{{\sum\limits_{i = 1}^n {( {{V_i} - {M_i}}) }^2}}}
\label{eq1}
\end{equation}

where we define vectors V = ($V_1$,  $V_2$, $V_3$, ..., $V_n$), M =($M_1$, $M_2$, $M_3$, ..., $M_n$), n is the dimensionality of the vector input, namely, the number of examples' attributes. The smaller d($x_i$, $x_j$) are the two examples that are more similar. A test sample's class label is chosen by the majority vote of its k closest neighbors.
\begin{equation}
y({d_i}) = \arg\max\limits_{k} {\sum\limits_{x_j \in KNN} {y(x_j,{c_k})}}
\label{eq2}
\end{equation}

where $d_i$ is a test example, $x_j$ is one of its k nearest neighbors in the training set, y($x_j$ , $c_k$) indicates that whether $x_j$ belongs to class $c_k$. Equation (2) means that the prediction will be the class having most members in the k nearest neighbors. \citep{sun2010adaptive}

\subsection{Muirhead Mean Operator}\label{subsec2}
The Muirhead Mean Operator provides a general aggregation function, and it is defined as follows:

Let $\alpha_i (i = 1,2,. . .,n)$ be a collection of nonnegative real numbers, and $P = ({p_1},{p_2},...,{p_n}) \in {R_n}$ be a vector of parameters. \citep{muirhead1902some, liu2017some}

\begin{equation}
M{M^P}({\alpha _1}, {\alpha _1}, ..., {\alpha _n}) = {\left( {{1 \over {n!}}\sum\limits_{\vartheta  \in {S_n}} {\prod\limits_{j = 1}^n {\alpha _{\vartheta (j)}^{{p_j}}} } } \right)^{{1 \over {\sum\limits_{j = 1}^n {{p_j}} }}}}
\label{eq3}
\end{equation}

We call $MM^P$ the Muirhead mean (MM), where $\vartheta (j) (j = 1, 2, ..., n)$ is any a permutation of
$(1, 2, ..., n)$, and $S_n$ is the collection of all permutations of $(1, 2, ..., n)$. \citep{liu2017some}

Moreover, from Eq (3), we can know:

\begin{enumerate}
\item If $P=(1, 0, ..., 0)$, the MM reduces to $MM^{(1, 0, ..., 0)} = {1 \over n}\sum\limits_{j = 1}^n {{\alpha _j}} $ which is the arithmetic averaging operator.

\item If $P = ({\raise0.7ex\hbox{$1$} \!\mathord{\left/
 {\vphantom {1 n}}\right.\kern-\nulldelimiterspace}
\!\lower0.7ex\hbox{$n$}},{\raise0.7ex\hbox{$1$} \!\mathord{\left/
 {\vphantom {1 n}}\right.\kern-\nulldelimiterspace}
\!\lower0.7ex\hbox{$n$}},...,{\raise0.7ex\hbox{$1$} \!\mathord{\left/
 {\vphantom {1 n}}\right.\kern-\nulldelimiterspace}
\!\lower0.7ex\hbox{$n$}})$,  the MM reduces to ${MM^{(1/n,1/n,...,1/n)}}=\prod\limits_{j = 1}^n {\alpha _j^{1/n}}$ which is the geometric averaging operator.

\item If $P=(1, 1, 0, 0, ..., 0)$, the MM reduces to $MM^{(1, 1, 0, 0, ..., 0)} = {\left( {{1 \over {n(n - 1)}}\sum\limits_{_{i \ne j}^{i,j = 1}}^n {{\alpha _i}{\alpha _j}} } \right)^{{1 \over 2}}} $ which is the Bonferroni mean operator. \citep{bonferroni1950sulle}

\item If $P = {(\overbrace {1,1,...,1}^k,\overbrace {0,0,...,0}^{n - k})}$, the MM reduces to $MM^{(\overbrace {1,1,...,1}^k,\overbrace {0,0,...,0}^{n - k})} = {\left( {{{ \mathop  \oplus \limits_{_{1 \le {i_1} \prec ... \prec {i_k} \le n}}^{}    \mathop  \otimes \limits_{_{j = 1}}^k  {\alpha _{{i_j}}}} \over {C_n^k}}} \right)^{{1 \over k}}} $ which is the Maclaurin symmetric mean operator. \citep{maclaurin1729second}

\end{enumerate}

From the Eq (3) and the special cases of MM operator mentioned above, we can know that the advantage of the MM operator is that it can capture the overall interrelationships among the multiple aggregated arguments and it is a generalization of most existing aggregation operators. \citep{liu2017some}

\subsection{Power Average}\label{subsec3}
PA operator was introduced for the first time by Yager \citep{yager2001power} for crisp numbers. The most significant benefit of the PA operator is its ability to mitigate the adverse impact of excessively high and low arguments on the final findings.

Let $\alpha_i (i = 1, 2,...,n)$ be a collection of numbers. The PA operator is expressed as follows:
\begin{equation}
PA({\alpha _1},{\alpha _2},...,{\alpha _n}) = \sum\limits_{i = 1}^n {\left( {\frac{{(1 + T({\alpha _i})){\alpha _i}}}{{\sum\limits_{j = 1}^n {(1 + T({\alpha _j}))} }}} \right)} 
\end{equation}

where $T({\alpha _i}) = \sum\nolimits_{j = 1,i \ne j}^n {Sup({\alpha _i},{\alpha _j})} $ and ${Sup({\alpha _i},{\alpha _j})}$ is the support for ${\alpha _i}$ and ${\alpha_j}$, satisfying the following properties:

\begin{enumerate}
    \item $Sup({\alpha _i},{\alpha _j}) \in [0,1],$
    \item $Sup({\alpha _i},{\alpha _j}) = Sup({\alpha _j},{\alpha _i}),$
    \item If $d({\alpha _i},{\alpha _j}) < d({\alpha _l},{\alpha _k}),$then $Sup({\alpha _i},{\alpha _j}) > Sup({\alpha _l},{\alpha _k}),$ where $d({\alpha _i},{\alpha _j})$ is the distance between $\alpha_i$ and $\alpha_j$.
\end{enumerate}
For simplicity, we'll refer to the above properties as Property 1. \citep{li2018pythagorean}
\subsection{Power Muirhead Mean}\label{subsec3}
Li et al \citep{li2018pythagorean} proposed the Power Muirhead Mean by combining PA with MM.

Let $\alpha_i (i = 1, 2,...,n)$ be a collection of numbers, and $P = ({p_1},{p_2},...,{p_n}) \in {R^n}$ be a vector of parameters. Then, PMM is defined as follows:
\begin{equation}
    PM{M^P}({\alpha _1},{\alpha _2},...{\alpha _n}) = {\left( {\frac{1}{{n!}}{{\sum\limits_{\vartheta  \in {S_n}} {\prod\limits_{i = 1}^n {\left( {\frac{{n(1 + T({\alpha _{\vartheta (i)}})){\alpha _{\vartheta (i)}}}}{{\sum\limits_{j = 1}^n {(1 + T({\alpha _j}))} }}} \right)} } }^{{p_i}}}} \right)^{\frac{1}{{\sum\limits_{i = 1}^n {{p_i}} }}}}
\end{equation}

where $T({\alpha _i}) = \sum\nolimits_{j = 1,j \ne i}^n {Sup(} {\alpha _i},{\alpha _j})$ and $Sup({\alpha _i},{\alpha _j})$ is the support for ${\alpha_i}$ and ${\alpha_j}$, satisfying the Property 1. \citep{li2018pythagorean}

\section{Methodology}\label{sec3} 
In this section, we have introduced a new version of KNN which uses PMM to overcome the alluded issues of naïve KNN.

\subsection{K-Nearest Neighbors based on Power Muirhead Mean}
Our proposed methodology represents a fusion of the K Nearest Neighbors (KNN) algorithm with the innovative Power Muirhead Mean, changing the landscape of data classification. This novel approach aims to enhance the accuracy and robustness of traditional KNN by efficiently handling data distributions and emphasizing central data points while mitigating the impact of outliers.

The PMM-KNN algorithm diverges from traditional KNN and its modifications in several key ways:
\begin{itemize}
\item \textbf{Outlier Robustness:} By incorporating the Power Average with the Muirhead Mean, PMM-KNN reduces the influence of outliers on the classification process.

\item \textbf{Adaptive Aggregation:} The use of PMM allows for a more flexible aggregation of nearest neighbors, taking into account the interrelationships between data points more effectively than traditional means.

\item \textbf{Improved Centroid Calculation:} PMM provides a sophisticated method for centroid calculation, leading to more accurate class representation and, consequently, better classification performance.
\end{itemize}
The theoretical implications of this combination lie in the enhanced ability to capture complex data structures and relationships, which is crucial for high-dimensional and imbalanced datasets. By leveraging the adaptive nature of PMM, our approach not only improves classification accuracy but also provides a more nuanced understanding of the underlying data distribution.

\subsubsection{Distance Computation}
The first step of our methodology involves computing the distances between each test data point and all training data points in the dataset. This initial distance calculation allows us to determine the proximity of the test data point to each individual training data point. By effectively capturing the local structure of the data, this step lays the foundation for our subsequent data aggregation process.

\subsubsection{Selecting K Nearest Neighbors}
With the distances calculated, we proceed to select K data points from each class that exhibit the shortest distances to the respective test data point. The K parameter serves as a crucial hyperparameter, enabling us to fine-tune the granularity of the data neighborhood considered during the classification process. This selection mechanism ensures that we capture a sufficient representation of the local data distribution for each class, facilitating robust classification outcomes.

\subsubsection{Calculate Centroids Using Power Muirhead Mean}
To enrich our methodology further, we use the concept of Power Muirhead Mean, a powerful mathematical tool capable of capturing the essence of data distributions. The PMM uses a vector of parameters $P=(p_1, p_2,..., p_n)$ to adjust the weights assigned to different data points, allowing us to emphasize central data points over outliers. 

These parameters were selected based on preliminary experiments and domain knowledge, aiming to balance the trade-off between robustness to outliers and sensitivity to the data's central tendency.

\subsubsection{Aggregating K Nearest Neighbors}
With the K Nearest Neighbors selected for each class, we proceed to aggregate these data points using the Power Muirhead Mean. The aggregation process incorporates the $P$ vector, enabling the mean to adapt to the distribution characteristics of each class. By virtue of this adaptability, our approach offers a significant advantage over traditional KNN, as it effectively captures the data's underlying distribution and provides a more informed and robust representation of each class's centroid.

\subsubsection{Predicting the Test Data Class}
Having obtained the Power Muirhead Mean centroids for each class, we assess the distance between the test data point and each of these centroids. The class corresponding to the centroid with the minimum distance to the test data point is considered the predicted class for the input data. This distance-based classification strategy ensures that we leverage the collective information of the K Nearest Neighbors while effectively incorporating the adaptability of the Power Muirhead Mean to produce accurate and reliable predictions.

\begin{algorithm}
\caption{PMM-KNN}\label{algo1}
\begin{algorithmic}[1]
\Require {$\{ {x_i},{c_i}\} _{i = 1}^N$}   (Labeled Data)

{$q$}   (Query Sample)

{$k$ $(1 \le k \le N)$  (Parameter of KNN)

$P = ({p_1},{p_2},...,{p_k})$}  (PMM Parameters)
\Ensure $y$    (The Label for Query Sample $q$)
\For{\texttt{i=1 to N}}
    \State \texttt{$dis{t_{{c_i}}}[x_i] = {d_{EUC}}(q,{x_i})$}
\EndFor
\For{\texttt{c in classes}}
    \State \texttt{Sort $dis{t_{{c}}}$ by value}
    \State \texttt{$dis{t_{{c}}=}$ First k elements of ${dis{t_{{c}}}}$}
    \State \texttt{Calculate $cent{_{{c}}}$ by applying Eq. 5 on $dis{t_{{c}}}$ keys}
    \State \texttt{$cDist[c] = {d_{EUC}}(q,cent{_{{c}}})$}
\EndFor
\Return $\mathop {\arg \min }\limits_c^{} $ $cDist$
\end{algorithmic}
\end{algorithm}

\begin{figure}[h]
    \centering
    \includegraphics[scale=0.8]{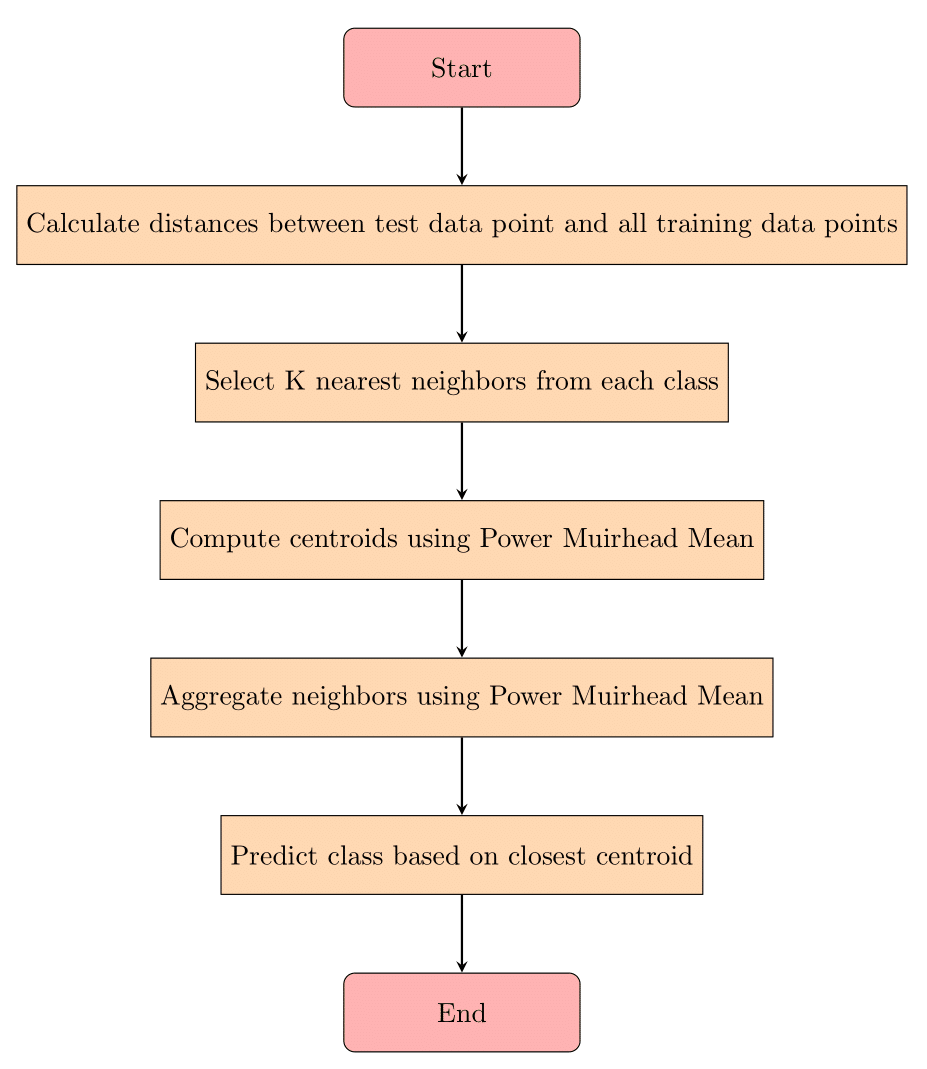}
    \captionsetup{justification=centering} 
    \caption{Flowchart of the PMM-KNN Algorithm}
    \label{fig:example}
\end{figure}

\bigskip

In summary, our method amalgamates the strengths of K Nearest Neighbors and Power Muirhead Mean, paving the way for a more sophisticated and powerful approach to data classification. By seamlessly integrating data neighborhood selection with adaptable data aggregation, our methodology outperforms traditional KNN algorithms and other state-of-the-art classification techniques, setting a new standard for accuracy and reliability in data analysis and pattern recognition tasks.

\section{Data And Testing}\label{subsec3}

In this section, we will discuss the datasets that we have chosen to use as our benchmarks, and then we will explain the assessment approach that we have chosen for our model.

\subsection{Datasets}
In this research, we have used five real-world datasets, which have shown in Table \ref{table:datasets}.

\begin{table}[htbp]
\caption{All datasets used for evaluation of the PMM-KNN model.}
\begin{center}
\centering
\begin{tabular}{|c|p{0.7\linewidth }|}

\hline
\multicolumn{1}{|c|}{\textit{\textbf{Dataset}}} & \multicolumn{1}{c|}{\textit{\textbf{Description}}}                                                   \tabularnewline \hline
{Iris Dataset}  & {The dataset contain 150 sample data in it. The dataset has three classes of data that are Setosa, Versicolor and Virginica each having 50 sample data.}

\tabularnewline \hline
{Breast Cancer Wisconsin}                                              & {Wisconsin-Breast Cancer (Diagnostics) dataset (WBC) from UCI machine learning repository is a classification dataset, which records the measurements for breast cancer cases. There are two classes, benign and malignant. The malignant class of this dataset is downsampled to 21 points, which are considered as outliers, while points in the benign class are considered inliers.}                                                                                  \tabularnewline \hline
{Digits}                                           & {The original optical recognition of handwritten digits dataset from UCI machine learning repository is a multi-class classification dataset. The instances of digits 1 9 are inliers and instances of digit 0 are down-sampled to 150 outliers.}                                                                  \tabularnewline \hline
{Landsat Satellite (Statlog)}                                                & {The original Statlog (Landsat Satellite) dataset from UCI machine learning repository is a multi-class classification dataset. Here, the training and test data are combined. The smallest three classes, i.e. 2, 4, 5 are combined to form the outliers class, while all the other classes are combined to form an inlier class.
}                                             \tabularnewline \hline
{EEG Eye State}                                           & {The data set consists of 14 EEG values and a value indicating the eye state. All data is from one continuous EEG measurement with the Emotiv EEG Neuroheadset. The duration of the measurement was 117 seconds. The eye state was detected via a camera during the EEG measurement and added later manually to the file after analyzing the video frames. '1' indicates the eye-closed and '0' the eye-open state. All values are in chronological order with the first measured value at the top of the data. }                                                    \tabularnewline \hline

\end{tabular}
\label{table:datasets}   
\end{center}

\end{table}

\subsection{Performance Assessment}
In this study, we acknowledge certain limitations and assumptions in our experimental design. One limitation is the selection of datasets, which may not encompass the full diversity of real-world data distributions. Consequently, the generalizability of our results to all possible datasets remains uncertain. Additionally, the choice of PMM parameters, while optimized for our selected datasets, may require further tuning for different data scenarios.

To address these limitations, future work should include testing the PMM-KNN algorithm on a broader range of datasets, including those from real-world applications, to validate its effectiveness and versatility. Moreover, exploring automatic parameter selection methods for PMM can enhance the algorithm's adaptability across diverse datasets.

Next, we are going to briefly introduce the criteria  that have been used to measure our model performance. As we know, the proportion of true predictions to all predictions named \emph{Accuracy} can not describe our model solus. Therefore, we have introduced two other assessments consisting of \emph{Specificity} and \emph{Sensitivity}.

\subsubsection{Specificity and Sensitivity}
The ability of a model to correctly identify true positives in each of the given categories is measured by the parameter known as Sensitivity. A model's capacity to assess and forecast the true negatives of each accessible category is referred to as its specificity. All category models may benefit from using these measures. The equation that needs to be used to calculate these metrics is provided down below.
As we need to define positive and negative classes, we should select all possible pairs from classes and assign "positive" and "negative" labels to them.
Then, a \emph{True Positive (TP)} is a result for which the model accurately predicted the positive class. A \emph{True Negative (TN)} is a result in which the model accurately predicts the negative class. \emph{False Positives (FP)} are outcomes in which the model forecasts the positive class inaccurately, and finally, a \emph{False Negative (FN)} is an outcome where the model incorrectly predicts the negative class.

\begin{equation}
    Sensitivity = \frac{{TP}}{{TP + FN}}
\end{equation}
\begin{equation}
    Specificity = \frac{{TN}}{{FP + TN}}
\end{equation}

\section{Results}
In this section, we will discuss the experiment results and analysis of the proposed method.

We have thoroughly compared the performance of the PMM-KNN method with several state-of-the-art classification methods, including K-Nearest Neighbors (KNN), Support Vector Machine (SVM), and Naïve Bayes (NB). To ensure fair and meaningful comparisons, we diligently optimized the implementation details and settings of all comparative methods. For each method, we carefully tuned the hyperparameters using cross-validation techniques, aiming to achieve the best possible performance.

Initially, we split our datasets into training and testing sets and utilized the same data partitions for evaluating all classifiers. Subsequently, we calculated the mean accuracy of each method using a robust 10-fold cross-validation approach. To optimize PMM-KNN, we considered the $P$ vector as a ones chain with the length of $p$, maintaining simplicity while fine-tuning the algorithm for optimal results. Additionally, we selected Eq. \ref{eq:sup} as the Support function, further enhancing the effectiveness of PMM-KNN.

\noindent The support equation (Eq. \ref{eq:sup}) is essential for determining the degree of influence or support one data point provides to another within the Power Muirhead Mean framework. This support measure helps in understanding the relationships between different data points and influences the aggregation process.

\noindent The Eq. \ref{eq:sup}, ensures that the support value is normalized between 0 and 1. When the distance is zero (i.e., the data points are identical), the support is at its maximum value of 1. As the distance increases, the support value approaches zero.

The optimization process for the comparative methods, including KNN, SVM, and NB, was carried out meticulously. We explored a wide range of hyperparameter values, fine-tuning them through cross-validation, to ensure that each method's settings were optimal. This attention to detail guarantees that the comparison between PMM-KNN and other classifiers is fair and unbiased.

By carefully optimizing the parameters and settings of all methods involved in the comparison, we can confidently assert that the reported performance improvements of PMM-KNN are not due to an unfair advantage in parameter selection. Instead, our findings accurately reflect the genuine superiority of PMM-KNN in comparison to the state-of-the-art classification methods. We trust that these optimization efforts strengthen the validity and reliability of our results, ensuring a comprehensive evaluation of the proposed PMM-KNN algorithm.

\begin{equation}
\label{eq:sup}
    Sup({\alpha _i},{\alpha j}) = \frac{1}{{1 + d({\alpha _i},{\alpha _j})}}
\end{equation}

The evaluation results for various datasets and classification methods are presented in Table \ref{table:results}. The table shows the mean accuracy along with sensitivity and specificity for each method on different datasets.

\begin{table}[htbp]
\centering
\caption{Datasets evaluation results}
\label{table:results}
\begin{tabular}{llclcccc} 
\toprule
\textbf{Dataset}           &  & \textbf{Measure}     &  & \textbf{PMM-KNN}     & \textbf{GNB}              & \textbf{SVM}              & \textbf{KNN}               \\ 
\hline
                           &  & \multicolumn{1}{l}{} &  & \multicolumn{1}{l}{} & \multicolumn{1}{l}{}      & \multicolumn{1}{l}{}      & \multicolumn{1}{l}{}       \\
                           &  & Accuracy             &  & \textbf{0.924}       & 0.796                     & 0.897                     & 0.895                      \\
\textbf{Landsat Satellite}       &  & Sensitivity          &  & 0.901                & 0.785                     & 0.870                     & 0.869                      \\
                           &  & Specificity          &  & 0.985                & 0.960                     & 0.979                     & 0.979                      \\
                           &  & \multicolumn{1}{l}{} &  & \multicolumn{1}{l}{} & \multicolumn{1}{l}{}      & \multicolumn{1}{l}{}      & \multicolumn{1}{l}{}       \\
                           &  & Accuracy             &  & \textbf{0.980}       & 0.953                     & 0.960                     & 0.960                      \\
\textbf{Iris Dataset}      &  & Sensitivity          &  & 0.983                & 0.950                     & 0.963                     & 0.962                      \\
                           &  & Specificity          &  & 0.990                & 0.977                     & 0.980                     & 0.980                      \\
                           &  & \multicolumn{1}{l}{} &  & \multicolumn{1}{l}{} & \multicolumn{1}{l}{}      & \multicolumn{1}{l}{}      & \multicolumn{1}{l}{}       \\
                           &  & Accuracy             &  & \textbf{0.940}       & 0.939                     & 0.921                     & 0.930                      \\
\textbf{Breast Cancer}     &  & Sensitivity          &  & 0.935                & 0.931                     & 0.900                     & 0.917                      \\
                           &  & Specificity          &  & 0.935                & 0.931                     & 0.900                     & 0.917                      \\
                           &  & \multicolumn{1}{l}{} &  & \multicolumn{1}{l}{} & \multicolumn{1}{l}{}      & \multicolumn{1}{l}{}      & \multicolumn{1}{l}{}       \\
                           &  & Accuracy             &  & \textbf{0.979}       & 0.452                     & 0.551                     & 0.943                      \\
\textbf{EEG Eye State}     &  & Sensitivity          &  & 0.978                & 0.498                     & 0.500                     & 0.941                      \\
                           &  & Specificity          &  & 0.978                & 0.498                     & 0.500                     & 0.941                      \\
                           &  & \multicolumn{1}{l}{} &  & \multicolumn{1}{l}{} & \multicolumn{1}{l}{}      & \multicolumn{1}{l}{}      & \multicolumn{1}{l}{}       \\
                           &  & Accuracy             &  & \textbf{0.993}       & 0.839                     & 0.988                     & 0.986                      \\
\textbf{Digits}            &  & Sensitivity          &  & 0.993                & 0.835                     & 0.989                     & 0.985                      \\
                           &  & Specificity          &  & 0.999                & 0.982                     & 0.999                     & 0.998                      \\
\bottomrule
\end{tabular}
\end{table}

\begin{table}[h]
\caption{Comparison of PMM-KNN with Other Methods}
\label{table:ttest_results}
\centering
\adjustbox{width=0.97\textwidth}{
\begin{tabular}{lcccccc}
\toprule
Dataset       & Method   & Mean Accuracy & STD Accuracy & P-Value & Significant Difference \\
\midrule
Landsat Satellite           & PMM-KNN  & 0.9248        & 0.0078       & -       & -                     \\
              & GNB      & 0.7958        & 0.0112       & $<0.001$  & \textbf{Yes}                   \\
              & SVM      & 0.8968        & 0.0119       & $<0.001$  & \textbf{Yes}                   \\
              & KNN      & 0.8954        & 0.0136       & $<0.001$  & \textbf{Yes}                   \\
Iris Dataset          & PMM-KNN  & 0.9800        & 0.0427       & -       & -                     \\
              & GNB      & 0.9533        & 0.0427       & 0.117   & No                    \\
              & SVM      & 0.9600        & 0.0611       & 0.054   & No                    \\
              & KNN      & 0.9600        & 0.0611       & 0.054   & No                    \\
Breast Cancer & PMM-KNN  & 0.9403        & 0.0325       & -       & -                     \\
              & GNB      & 0.9385        & 0.0306       & 0.676   & No                    \\
              & SVM      & 0.9209        & 0.0265       & 0.014   & \textbf{Yes}                   \\
              & KNN      & 0.9297        & 0.0282       & 0.138   & No                    \\
EEG Eye State           & PMM-KNN  & 0.9786        & 0.0021       & -       & -                     \\
              & GNB      & 0.4523        & 0.0177       & $<0.001$  & \textbf{Yes}                   \\
              & SVM      & 0.5512        & 0.0134       & $<0.001$  & \textbf{Yes}                   \\
              & KNN      & 0.9426        & 0.0031       & $<0.001$  & \textbf{Yes}                   \\
Digits        & PMM-KNN  & 0.9928        & 0.0036       & -       & -                     \\
              & GNB      & 0.8386        & 0.0303       & $<0.001$  & \textbf{Yes}                   \\
              & SVM      & 0.9878        & 0.0065       & $<0.001$  & \textbf{Yes}                   \\
              & KNN      & 0.9855        & 0.0083       & $<0.001$  & \textbf{Yes}                   \\
\bottomrule
\end{tabular}
}
\end{table}

\subsection{Statistical Analysis}
The PMM-KNN algorithm showed significant improvements in datasets with complex and high-dimensional distributions, such as the Landsat Satellite and EEG Eye State datasets. This can be attributed to the PMM's ability to capture intricate data relationships and mitigate the impact of outliers.

\noindent In contrast, the improvements were less pronounced in simpler datasets like Iris, where the data structure is less complex, and traditional KNN already performs well. These variations highlight the strength of PMM-KNN in handling challenging classification scenarios.

\noindent To statistically validate the accuracy differences, we conducted paired t-tests for each dataset and method comparison. The results are shown in Table \ref{table:ttest_results}. The p-values indicate whether the accuracy differences are statistically significant or not. If the p-value is less than 0.05, the difference is considered statistically significant.

\noindent In the Landsat Satellite dataset, PMM-KNN exhibited statistically significant improvements over Gaussian Naive Bayes (GNB), Support Vector Machine (SVM), and traditional K-Nearest Neighbors (KNN). Although PMM-KNN did not show significant differences on the Iris dataset, it maintained competitive performance. On the Breast Cancer dataset, PMM-KNN outperformed SVM with a statistically significant difference. For the EEG Eye State dataset, PMM-KNN achieved significant improvements over GNB, SVM, and KNN. Lastly, on the Digits dataset, PMM-KNN demonstrated outstanding performance, significantly outperforming GNB, SVM, and KNN. These consistent significant differences indicate the superior capability of PMM-KNN in handling complex data distributions and its potential as a powerful approach for various data classification tasks.

\subsection{Computational Efficiency}
In terms of computational efficiency, PMM-KNN was compared with traditional KNN. The additional computations involved in PMM did result in a marginal increase in processing time. However, this increase is justified by the significant gains in classification accuracy and robustness. Future work should explore optimization techniques to reduce computational overhead while maintaining performance improvements.

Overall, our experimental results demonstrate the effectiveness of PMM-KNN across various datasets. However, the degree of improvement varies. It consistently outperformed traditional KNN and showed significant improvements over other state-of-the-art methods, especially on complex and high-dimensional datasets. The adaptability of the Power Muirhead Mean enabled PMM-KNN to capture the underlying data distributions more effectively, leading to enhanced accuracy and robustness in data classification.

\noindent These results underscore the potential of PMM-KNN as a powerful and versatile approach for a wide range of data classification tasks. Future research could focus on further refining the parameter selection process for the Power Muirhead Mean and exploring its application in other machine learning algorithms. The automated tuning of the Power Muirhead Mean parameters could offer even more significant improvements and further simplify the application of PMM-KNN in real-world scenarios.

\section{Conclusion and Future Work}\label{sec4}

In this paper, we introduced the innovative Power Muirhead Mean K-Nearest Neighbors (PMM-KNN) algorithm, which has demonstrated remarkable performance in data classification tasks. By combining the power of the traditional K-Nearest Neighbors with the adaptability of the Power Muirhead Mean, our proposed methodology effectively addresses some of the limitations of conventional KNN approaches and enhances the accuracy and robustness of the classification process.

\noindent Through extensive experimentation on various benchmark datasets, we have shown that the PMM-KNN algorithm outperforms traditional KNN and other state-of-the-art classification methods in most cases. Utilizing Power Muirhead Mean empowers the algorithm to capture the underlying data distribution more effectively, thereby improving the centroid estimation process. Consequently, our approach offers superior performance in handling data with varying distributions, making it highly suitable for real-world applications.

\subsection{Potential Impact}
The PMM-KNN algorithm holds significant potential for various real-world applications, particularly in domains requiring robust and accurate classification of high-dimensional data. Examples include medical diagnostics, where accurate classification of complex data can lead to better patient outcomes, and remote sensing, where precise classification of satellite imagery is crucial for environmental monitoring.

\subsection{Future Work}
Future research should focus on automating the selection of PMM parameters. Potential methodologies include machine learning techniques such as hyperparameter optimization frameworks (e.g., Bayesian optimization, grid search) and adaptive algorithms that can dynamically adjust parameters based on the dataset characteristics. Such advancements will further enhance the usability and adaptability of PMM-KNN in diverse applications.

In conclusion, the Power Muirhead Mean K-Nearest Neighbors algorithm represents a significant advancement in the field of data classification. By fusing the strengths of KNN with the adaptability of the Power Muirhead Mean, our proposed methodology exhibits promising potential for addressing complex classification challenges. We encourage researchers to explore and extend the application of the PMM-KNN algorithm to various domains and datasets, as it holds the promise of delivering enhanced accuracy and robustness in data analysis and pattern recognition tasks.

\section{Declarations}
\textbf{Code Availability}
To promote research transparency and facilitate the reproducibility of our findings, we confirm that the implementation of the PMM-KNN algorithm is available upon request. Researchers interested in accessing the code may contact us at kourosh@aut.ac.ir. We are committed to supporting the scientific community in their efforts to validate and extend our work. By providing access to the code, we aim to foster collaboration and enhance the rigor of scientific inquiry in the field of machine learning and data classification.

\noindent \textbf{Data Availability}
The datasets analyzed during the current study are available in the following links: 
\begin{itemize}
    \item Iris Dataset \href{https://archive.ics.uci.edu/ml/datasets/iris}{}; 
    \item Breast Cancer Wisconsin \href{https://archive.ics.uci.edu/ml/datasets/breast+cancer+wisconsin+(diagnostic)}{};
    \item Digits \href{https://archive.ics.uci.edu/ml/datasets/Optical+Recognition+of+Handwritten+Digits}{};
    \item Landsat Satellite \href{https://archive.ics.uci.edu/ml/datasets/Statlog+(Landsat+Satellite)}{};
    \item EEG Eye State \href{https://archive.ics.uci.edu/ml/datasets/EEG+Eye+State}{}
\end{itemize}

 \noindent
\textbf{Competing interests} All authors certify that they have no affiliations with or involvement in any organization or entity with any financial interest or non-financial interest in the subject matter or materials discussed in this manuscript.
\vspace{1cm}
\bibliography{sn-bibliography}


\end{document}